\documentclass{article}
\usepackage{nips07submit_e,times}
\usepackage{amssymb}
\usepackage{amsmath}
\usepackage{graphicx}

\title{Multi-Way, Multi-View Learning}
\author{
Ilkka  Huopaniemi and Tommi Suvitaival \\
Department of Information and Computer Science\\
Helsinki University of Technology\\
Finland\\
\texttt{ilkka.huopaniemi@tkk.fi}\\
\texttt{tommi.suvitaival@tkk.fi}\\
\And
Janne Nikkil\"a  \\
Department of Basic Veterinary Sciences\\ 
Faculty of Veterinary Medicine\\ 
University of Helsinki\\
Finland  \\
\texttt{janne.nikkila@helsinki.fi}\\
\And
Matej Ore\v{s}i\v{c}\\
VTT Technical Research Centre of Finland\\
Espoo, Finland\\
\texttt{matej.oresic@vtt.fi}\\
\And
Samuel Kaski \\
Dept of Information and Computer Science\\
Helsinki University of Technology\\
Finland\\
\texttt{samuel.kaski@tkk.fi}
}
\begin{document}
\maketitle

\begin{abstract}
  We extend multi-way, multivariate ANOVA-type analysis to cases where
  one covariate is the view, with features of each view coming from
  different, high-dimensional domains. The different views are assumed
  to be connected by having paired samples; this is a common setup in
  recent bioinformatics experiments, of which we analyze metabolite
  profiles in different conditions (disease vs.\ control and treatment
  vs.\ untreated) in different tissues (views). We introduce a multi-way
  latent variable model for this new task, by extending the generative
  model of Bayesian canonical correlation analysis (CCA) both to take
  multi-way covariate information into account as population priors,
  and by reducing the dimensionality by an integrated factor analysis
  that assumes the metabolites to come in correlated groups.
\end{abstract}

\section{Introduction}

Finding disease and treatment effects from populations of measurements
is a prototypical multi-way modeling task, traditionally solved with
multivariate ANOVA. Here disease state (diseased/healthy) and
treatment (treated/placebo) are the two covariates, and the research question
is, are there differences in the population that can be explained by
either covariate or, more interestingly, their interaction, which
would hint at the treatment being effective. It is naturally
additionally interesting what the differences are.

A recurring problem in multi-way analyses, especially with modern
high-throughput measurements in molecular biology, is the ''small $n$,
large $p$''-problem. The dimensionality $p$ of the measurements is
high while the number of samples $n$ is low, and additionally the data
may be collinear making estimation of the effects impossible with
classical methods, univariate or multivariate linear models solved
with multi-way ANOVA techniques. The most promising modern method,
Bayesian sparse factor regression model \cite{West03}, is useful in
finding the variables most strongly related to the external covariate
and to infer relationships between those variables via common latent
factors. Instead of a regression model we will use a generative latent
factor model which incorporates an assumption of clusteredness of the
variables to regularize the model, and makes it possible to extend the
model to multi-view factor analysis. Such clusteredness is well
justified in our application field, metabolomics, where due to
biochemical reaction pathways the variation in
concentrations of metabolite groups is highly correlated \cite{Steuer06}.

Assume that measurements have been made on the same objects but with
different methods, resulting in different data sources possibly on
different domains. An example we will analyze in this paper is
metabolomic profiles in different tissues, where the domains are
partly different since the metabolites cannot be fully matched. The
different views form one covariate in the multi-way analysis, with
the additional problem that the samples come from different domains
and cannot be directly compared. We introduce a new hierarchy level of
latent variables intended to decompose the views into view-specific
and shared components, which is needed for the multi-way
analysis. Such a decomposition is possible given that the samples in
the different views come in pairs, which we need to assume.

The resulting decomposition between the views turns out to be
implementable with Bayesian canonical correlation analysis
\cite{Klami07icml,Wang07,Archambeau08}, interpretable as unsupervised
multi-view modeling. Hence, in this work we re-interpret unsupervised
multi-view modeling as one-way modeling of samples from different
domains, and combine it with multi-way modeling. Given that we
additionally can work under the \emph{large p, small n} conditions,
the model is expected to have widespread applicability in current molecular
biological measurements.

\section{Model}
\label{Section-model}

\subsection{Multi-way, multi-view}

We will generalize ANOVA to multi-view (multi-domain) analysis,
restricting to two covariates and two views for simplicity. Using ANOVA-style notation and assuming the views to be in
the same domain, the multivariate linear model for samples is
\begin{equation}
\label{ANOVAmodel}
{\bf
  v}_d=\boldsymbol{\alpha}_a+\boldsymbol{\beta}_b+(\boldsymbol{\alpha\beta})_{ab}
+\boldsymbol{\gamma}_d+(\boldsymbol{\alpha\gamma})_{ad}+(\boldsymbol{\beta\gamma})_{bd}+(\boldsymbol{\alpha\beta\gamma})_{abd}+\text{noise},
\end{equation}
where $a$ and $b$ ($a=0,\dots A$ and $b=0,\dots B$), are the two
traditional independent covariates such as disease and treatment, and
$d$ denotes the view. 

For different values of $d$ the domain of ${\bf v}^d$ may vary,
meaning different feature spaces with different dimensionalities. We
assume the samples of the different views to come in pairs,
$\mathbf{v}=[\mathbf{x},\mathbf{y}]$. For the rest of the paper we
will change the notation for clarity to $\mathbf{v}^1=\mathbf{x}$,
$\mathbf{v}^2=\mathbf{y}$, and assume a mapping $f^x$ from the effects
to the domain of $\mathbf{x}$ which is linear for now. Then,
\begin{equation}
{\bf
  x}=f^x(\boldsymbol{\alpha}_a+\boldsymbol{\beta}_b+(\boldsymbol{\alpha\beta})_{ab})
+f^x((\boldsymbol{\alpha})^x_{a}+(\boldsymbol{\beta})^x_{b}+(\boldsymbol{\alpha\beta})^x_{ab})+\text{noise},
\label{eq:threeway}
\end{equation}
assuming $\gamma_d=0$, because it does not make sense to compare means of different domains, and that the view-specific effects are in the
same domain as the view-independent effects and hence need to be
transformed with the same function. The equation for $\mathbf{y}$ is
analogous.  

To our knowledge, there exists no method capable of studying the view-independent, and view-dependent effects. In the next section we will
introduce a model which will additionally assume that the effects may
be uncertain, resulting in a hierarchical Bayesian model.

\subsection{Hierarchical model}

\begin{figure}
\begin{center}
\includegraphics[width=0.7\linewidth]{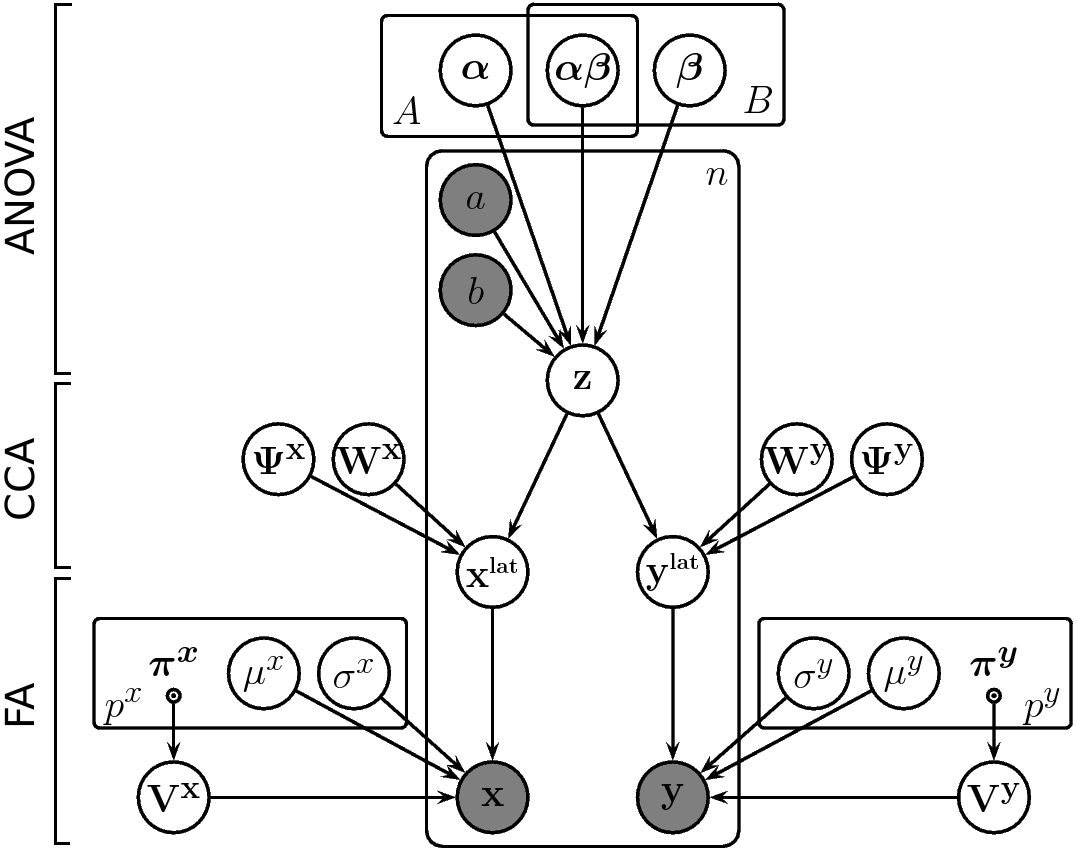}
\end{center}
\caption{The hierarchical latent-variable model for multi-way,
  multi-view learning under ``large $p$, small $n$'' conditions.}
\label{graphical_model}
\end{figure}

We next formulate a hierarchical latent-variable model for the task of
multi-way, multi-view learning under ``large $p$, small $n$''
conditions. For this we need three components: (i) regularized dimension reduction, (ii) combination of different data domains, and (iii) multi-way analysis. We formulate each of these as part of a big generative model. We will first summarize the main components of the model
shown in Figure~\ref{graphical_model}, and then describe each part in detail.

To deal with the small sample size $n \ll p$ problem, we reduce the
dimensionality of the data $\mathbf{x}$ and $\mathbf{y}$ from the two
views into their respective latent variables $\mathbf{x}^{lat}$ and
$\mathbf{y}^{lat}$. This is done with factor analyzers which assume
that the variables come in groups, which is a strongly regularizing
assumption effective under the ``large $p$, small $n$''
conditions. The clustering assumption is particularly sensible under
the assumption that metabolomics data, our main application, contains
strongly correlated groups of variables \cite{Steuer06}.

The second necessary element is search for a view shared by the two
different domains $\mathbf{x}$ and $\mathbf{y}$, needed for
finding shared multi-way effects. Given paired data, this is a task
for Bayesian CCA (BCCA) \cite{Klami07icml,Wang07,Archambeau08} which introduces a new hiearchy
level where a latent variable $\mathbf{z}$ captures the shared
variation between the views. The view-specific variation has been
implicitly modeled by view-specific latent variables which have been
integrated out, resulting in flexible covariance matrices
parameterized by the $\boldsymbol{\Psi}$. 

The third necessary element, the ANOVA-type two-way analysis is supplemented by assigning the effect terms as priors on the
latent variables $\mathbf{z}$; in normal BCCA the prior is
zero-mean. The observed covariates $a$ and $b$ choose the correct
effects for each sample. The covariates hence effectively change the
means of the data as in eqn (\ref{eq:threeway}), and the variation
around the mean is modeled with the rest of the model. The central
differences from (\ref{eq:threeway}) are that the model is
hierarchical, implying that the arguments of the linear function $f^x$
have a distribution, and that the ``noise'' is structured, stemming
from all the latent variables. With these additions, the model will be
better able to take into account the uncertainty in the data.

The posterior is computed with Gibbs sampling. The Gibbs-formulas are
included in the supplementary material.

In effect the model, shown in Figure~\ref{graphical_model}, consists
of two factor analyzers, where the loadings assume cluster memberships
(multiplied with scales), a generative model of CCA, and
population-specific priors on $\mathbf{z}$ that assume ANOVA-type
multi-way structure.  We will now introduce the details of each of these parts in
turn.

\subsubsection{Factor analysis model}
We need to reduce dimensionality, which can be done by factor analysis (FA).
The model \cite{Roweis99} for $n$ exchangeable
replicates is
\begin{eqnarray}
\mathbf{ x}_{j}^{lat} \sim \mathcal{N}(0,\boldsymbol{\Psi}^x) \nonumber \\
\mathbf{x}_j\sim \mathcal{N}({\boldsymbol{ \mu}^x}+\mathbf{V}^x \mathbf
{x}_{j}^{lat},\boldsymbol{ \Lambda}^x) \; .
\end{eqnarray}
Here $\boldsymbol{V}^x$ is the projection matrix that is
assumed to generate the data vector $\mathbf{x}_{j}$ from the latent
variable $\mathbf{x}_{j}^{lat}$. The $\mathbf{
x}_{j}^{lat}$ is a latent variable vector, whose elements are known
as factor scores.  The $\mathbf{V}^x \mathbf{x}^{lat}_j$ models
such common variance of the data around the variable-means
${\boldsymbol{\mu}^x}$ that can be explained by factors common to all or
many variables, effectively estimated from the sample covariance
matrix of the dataset. The sample covariance becomes decomposed into
$\hat{\boldsymbol{\Sigma}}=\mathbf{V}^x\mathbf{V}^{xT}
+\boldsymbol{\Lambda}^x$, where  ${\boldsymbol{ \Lambda}^x}$ is a diagonal residual
variance matrix with diagonal elements $\sigma_i^2$, modelling the
variable-specific noise not explained by the latent factors.  The covariance matrix of $\mathbf{x}^{lat}$, $\boldsymbol{\Psi}^x$, comes from the CCA.

At this point, when $n<p$, $\mathbf{V}^x$ cannot be estimated due to the
singularity of the sample covariance matrix. To overcome the $n \ll p$ problem,  we now restrict $\mathbf{V}^x$ to a non-singular clustering matrix, suitable for
data containing highly correlated groups of variables.

\subsubsection{Projection matrix that assumes grouped variables}

We make the structured assumption that there are strongly correlated
groups of variables in the data, the generated values within the whole group being
governed by one latent variable. The projection matrix
$\mathbf{V}^x$ is positive-valued, each row
having one non-zero element corresponding to the cluster assignment of
the variable,
\begin{equation}
\mathbf{V}^x=
\left[ \begin{array}{ccc}
\lambda_1 & 0 & 0 \\ 0 & 0 & \lambda_2\\
\vdots & \vdots & \vdots \\
0 & \lambda_j & 0 \\ 0& \lambda_{j+1} & 0 \\
\vdots & \vdots & \vdots \\
\end{array}  \right] \; .
\end{equation}
\label{Vmatrix}
The location of the non-zero value on row $i$, $v_i$, follows a multinomial distribution with one observation, with
an uninformative prior distribution $\boldsymbol{\pi}_i$. The $\boldsymbol{\pi}_i$ could also be used to encode prior information on the known grouping of
variables. The variation of each variable within a cluster is assumed to be
modeled by the same latent variable, but the scales $\lambda_i$ may
differ. The variable-specific residual variances $\sigma_i^2$, that are the
diagonal elements of $\boldsymbol{\Lambda}$, follow a scaled
$\textrm{Inv-} \chi^2$ with an uninformative prior.

In summary, we regularize the covariance matrix by assuming that the
main correlations are positive correlations between variables
belonging to the same cluster. This correlation is mediated through a
common latent variable; this is a reasonable assumption for
metabolomics data and, furthermore, facilitates interpretation of the results.

\subsubsection{Generative model of CCA}
We need to combine different data domains, and for paired data that can be done with CCA. The generative model of BCCA has been formulated
\cite{Klami07icml,Bach05} for sample $j$ as
\begin{align}
  \mathbf{z}_j & \sim N(0,\mathbf{I}), \notag \\
  \mathbf{x}^{lat}_j& \sim N(\mathbf{W}^x \mathbf{z}_j,
  \mathbf{\Psi}^x) \label{eq:PCCA},
\end{align}
and likewise for $\mathbf{y}$. Here we have assumed no mean parameter since the mean of the data is estimated in the factor analysis part. The $\mathbf{W}^x$ is a projection matrix from the latent
variables $\mathbf{z}_j$, and $\mathbf{\Psi}^x$ is a matrix of
marginal variances. The crucial thing is that the latent variables
$\mathbf{z}$ are shared between the two data sets, while everything
else is independent. The prior distributions were chosen as
\begin{align}
  \mathbf{w}_{l} & \sim N(0,\beta_l \mathbf{I}), \notag \\
  \beta_l & \sim IG(\alpha_0, \beta_0), \notag \\
  \mathbf{\Psi}^x,\mathbf{\Psi}^y & \sim IW(\mathbf{S}_0,\nu_0). \label{eq:prior}
\end{align}
Here $\mathbf{w}_l$ denotes the $l$th column of $\mathbf{W}$,
and $IG$ and $IW$ are shorthand notations for
the inverse Gamma and inverse Wishart distributions.
The priors for the covariance matrices $\mathbf{\Psi}^x$ and $\mathbf{\Psi}^y$ are conventional conjugate
priors, and the prior for the projection matrices is the so-called
Automatic Relevance Determination (ARD) prior used for example in
Bayesian principal component analysis \cite{Bishop99nips}.

\subsubsection{ANOVA-type model for  latent variables.}
\label{Model_ANOVA}

We assume that the ANOVA-type effects act on the latent variables
$\mathbf{z}$, which allows access to effects found in
both the spaces $\mathbf{x}^{lat}$ and $\mathbf{y}^{lat}$. They are
modeled as population priors to the latent variables, which in turn
are given Gaussian priors $\boldsymbol{ \alpha}_a$, $\boldsymbol{
  \beta}_b$, $(\boldsymbol{ \alpha\beta})_{ab} \sim
\mathcal{N}(0,\mathbf{I})$.

In the $K_z$-dimensional latent variable space we then have
\begin{equation} \mathbf{z}_j =\boldsymbol{
    \alpha}_a+ \boldsymbol{ \beta}_b+(\boldsymbol{ \alpha\beta})_{ab}
  +\boldsymbol{ \epsilon}_j,
\end{equation}
where $\boldsymbol{ \epsilon}_j$ is a noise term. 
Note that the grand means are estimated in the lower level of hierarchy,
that is, directly in the $\mathbf{x}$ and $\mathbf{y}$-spaces, and
do not appear here.

To simplify the interpretation of the effects we center the grand
means to the mean of one control population. A similar choice has been
done successfully in other ANOVA studies \cite{Seo07}, and it does not
significantly sacrifice generality. We set the parameter vector
$\boldsymbol{\mu}^x$, describing variable-specific means, to the mean
of the control group. One group now becomes
the baseline to which other classes are compared by adding main and
interaction effects. For convenience, we will additionally change the
variables compared to the standard ANOVA convention, such that the
terms $\boldsymbol{\alpha}_0$, $\boldsymbol{\beta}_0$,
$\boldsymbol{\alpha\beta}_{00}$, $(\boldsymbol{\alpha \beta})_{0b}$,
and $(\boldsymbol{\alpha \beta})_{a0}$ are not estimated. The
differences between the populations are now modelled directly with
$\mathbf{x}^{lat}$ and $\mathbf{y}^{lat}$, and hierarchically by the
main effects $\boldsymbol{\alpha}_a$, $\boldsymbol{\beta}_b$,
$(\boldsymbol{\alpha \beta})_{ab}$, $a,b>0$.

In our case study, $a$ and $b$ have only two values and we have populations $(a,b)={(0,0), (1,0),(0,1),(1,1) }$, and there are hence three terms $\boldsymbol{\alpha}_1$, $\boldsymbol{\beta}_1$ and $(\boldsymbol{\alpha\beta})_{11}$, that model the difference to the control population  $(a,b)=(0,0)$.

In summary, the complete hierarchical model of Figure~\ref{graphical_model} is 
\begin{eqnarray}
\boldsymbol{\alpha}_0=0,\boldsymbol{\beta}_0=0,(\boldsymbol{\alpha\beta})_{a0}=0,(\boldsymbol{\alpha\beta})_{0b}=0 \nonumber \\
\boldsymbol{\alpha}_a,\boldsymbol{\beta}_b,(\boldsymbol{\alpha\beta})_{ab}\sim \mathcal{N}(0,\mathbf{I}) \nonumber \\
{\mathbf{ z}}_{j}|_{j \in a,b} \sim N(\boldsymbol{\alpha}_a+\boldsymbol{\beta}_b +(\boldsymbol{\alpha\beta})_{ab},\mathbf{I}) \nonumber \\
{\mathbf{ x}}_{j}^{lat}\sim N(\mathbf{W}^x \mathbf{z}_j,\boldsymbol{\Psi}^x) \nonumber \\
{\mathbf{x}}_j\sim \mathcal{N}({\boldsymbol \mu}^x+{\mathbf{V}} \mathbf{x}_{j}^{lat},{\boldsymbol{ \Lambda}^x}). 
\end{eqnarray}

\subsection{Decomposing covariate effects into shared and view-specific}

\begin{figure}
\begin{center}
\includegraphics[width=0.6\linewidth]{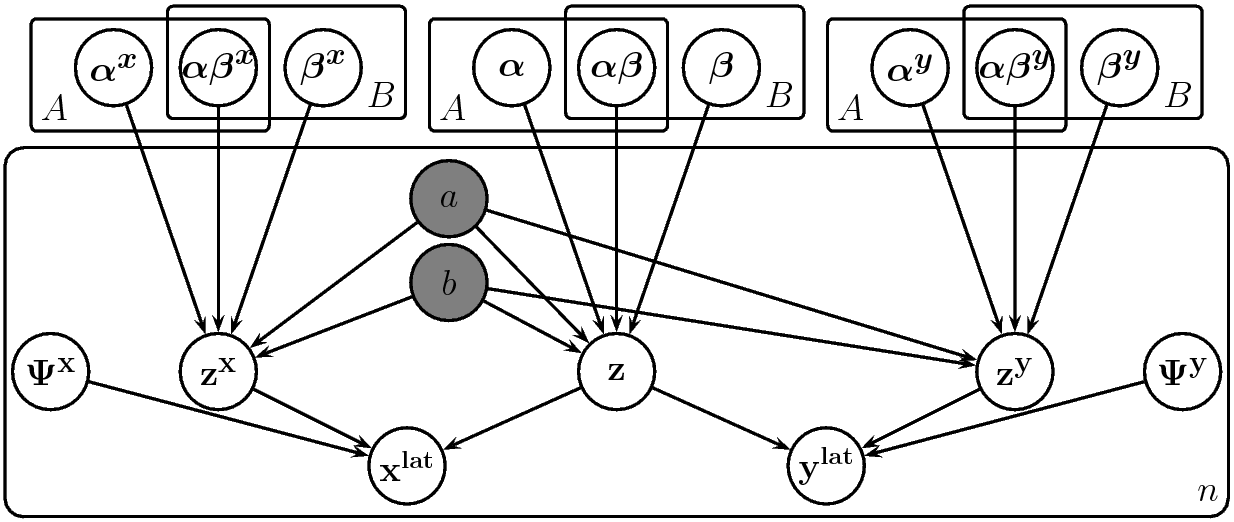}
\end{center}
\caption{The graphical model describing the decomposition of covariate
  effects into shared and view-specific ones. The figure expands
  the top part of Figure~\ref{graphical_model}, leaving out the
  feature extraction part and some parameters.}
\label{graphical_models_z}
\end{figure}

So far we have not discussed how the model finds the view-related
effects or, in our application, tissue effects
$\boldsymbol{\alpha}_a^x$, $\boldsymbol{\beta}_b^x$, and
$(\boldsymbol{\alpha\beta})_{ab}^x$, and likewise for $y$.

The Bayesian CCA assumes that the data is generated by a sum of
view-specific latent variables $\mathbf{z}^x$ and $\mathbf{z}^y$, and
shared latent variables $\mathbf{z}$, and the former have been
integrated out in the graphical model of
Figure~\ref{graphical_model}. The way to implement the view-specific
effects is to assign them as priors to the view-specific latent
variables. Then we do not want to integrate them out but include them
explicitly in the model as shown in Figure~\ref{graphical_models_z}.

As a technical note, to make computation of the model faster and more
reliable, we have further included view-specific latent variables that
do not have disease or treatment effects. They have been integrated
out, resulting in the covariance parameters $\boldsymbol{\Psi}$ in
Figure~\ref{graphical_models_z}. Their role is to explain away all or most of the variation
that is unrelated to the disease and treatment effects, so that Gibbs
sampling does not need to model all that variation. This trick should
not change modeling results in the limit of an infinite time for
computation.

In practice the decomposition in Figure~\ref{graphical_models_z} is
implemented by restricting a column of $\mathbf{W}^x$ to be zero for
the y-specific components and vice versa for x.

\subsection{Data preprocessing and model complexity selection}
For simplicity and to reduce the number of parameters of the model,
the data is preprocessed such that for each variable the mean of the
control population $a=0$, $b=0$ is subtracted and the variable is scaled by
the standard deviation of the control population. This fixes the scales
$\lambda_i$ to one and the $\boldsymbol{\mu}^x$ and $\boldsymbol{\mu}^y$ to zero. The factor analysis part now models
correlations of the variables. The possible covariate effects are now
comparable to the control population as discussed in Chapter
\ref{Model_ANOVA}

Model complexity, that is, the number of clusters and latent variables,
is chosen separately for both $\mathbf{x}^{lat}$ and
$\mathbf{y}^{lat}$ by predictive likelihood in 10-fold
cross-validation.


\section{Results}
We demonstrate the working of the method on generated data, and apply it to a disease study where lipidomic profiles have been measured from several tissues
of model mouse samples, under a two-way experimental setup (disease and treatment), the two feature spaces (lipid profiles) are distinct and
samples paired. 

\subsection{Generated data}

We generate data having known effects, and then study how well the model finds the effects as a function of the number of measurements. There are three effects, in $\alpha$, $\beta^y$, and $(\alpha\beta)^x$.

Each of the three effects have strength
$+2$, the $\mathbf{x}^{lat}$ and $\mathbf{y}^{lat}$ are both
3-dimensional, and the $\mathbf{x}$ and $\mathbf{y}$ are 200-dimensional. The
$\sigma_i=1$ for each variable $i$ in $\mathbf{x}$ and $\mathbf{y}$. The model is computed by Gibbs sampling,
discarding 1000 burn-in samples, and collecting 1000 samples for inference. To fix the sign of the effects without affecting the results, each
posterior distribution is mirrored, if necessary, to have a positive mean,
i.e. multiplied by the sign of the posterior mean.

\begin{figure}
\begin{center}
\includegraphics[width=0.7\linewidth]{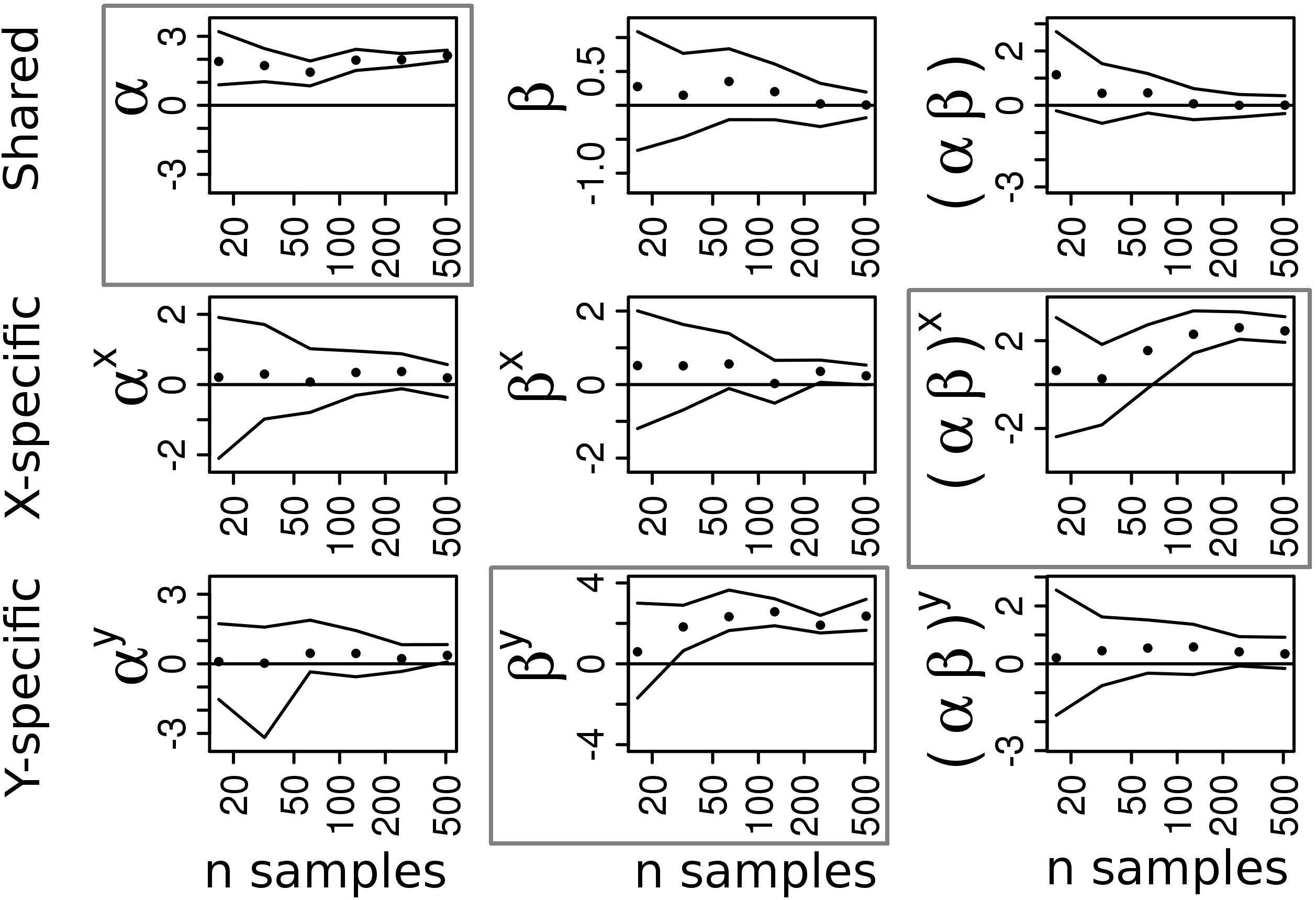}
\end{center}
\caption{The method finds the generated effects $\alpha=+2$,$\beta^y=+2$, and $(\alpha\beta)^x=+2$. (effect subscripts $ _1$ and $_{11}$ have been dropped in the rest of the results section). The dots show posterior mean and the thin lines include 95\% of posterior mass, as a function of number of observations. A consistently non-zero posterior distribution implies an effect found. }
\label{generated_data}
\end{figure}

The method finds the three generated effects, shown in Figure~\ref{generated_data}. The uncertainty decreases with increasing number of observations. The shared effect is found  with much less uncertainty since there is evidence from both views.  With low numbers of samples, there is considerable uncertainty in the effects for view-specific components. In typical bioinformatics applications there may be 20-50 samples.

\subsection{Lung cancer study}
We then study data from a two-way, two-view, $n \ll p$, so far
unpublished lung cancer mouse model experiment. The diseased mice are
compared to healthy control samples and, in addition, some mice from
both groups have been given a test anticancer drug treatment. There are thus
healthy untreated (9 mice/samples), diseased untreated (7), healthy
treated (6) and diseased treated (6) samples. Lipidomic profiles have
been measured by Liquid Chromatography Mass Spectrometry. The study
has a two-way experimental setup, such that disease effect $\alpha$,
treatment effect $\beta$ and an interaction effect $\alpha\beta$ on
lipid groups are to be estimated. The high-dimensional lipidomic
profiles have been measured from several tissues of each mouse; the
tissues have partly different lipids that have not been matched, and even the roles of the matched lipids may be different in different tissues. Hence, the tissues have
different feature spaces with paired samples, implying a two-view
study. We will specifically study the relationship between blood and lung tissue, which is the most interesting for diagnosis, since blood can be easily sampled.
 
\subsubsection{Experiment 1. Effects shared by blood and disease tissue}

\begin{figure}
\begin{center}
\begin{tabular}{cccc}
\includegraphics[width=0.0226\linewidth]{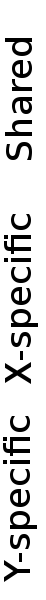}
\includegraphics[width=0.12\linewidth]{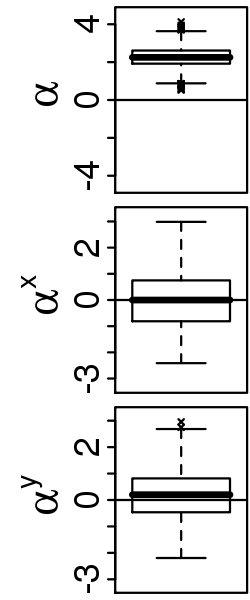}
\includegraphics[width=0.12\linewidth]{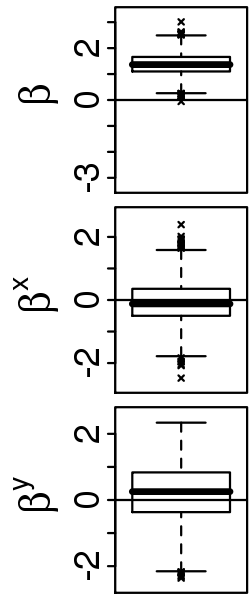}
\includegraphics[width=0.12\linewidth]{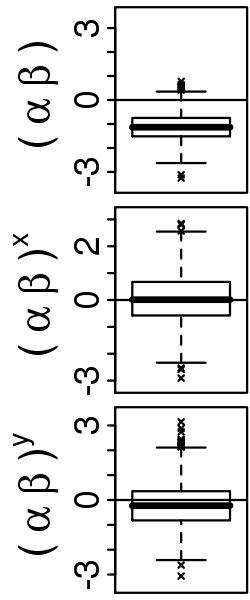}&
&  &

\includegraphics[width=0.0226\linewidth]{labels-realData.png}
\includegraphics[width=0.12\linewidth]{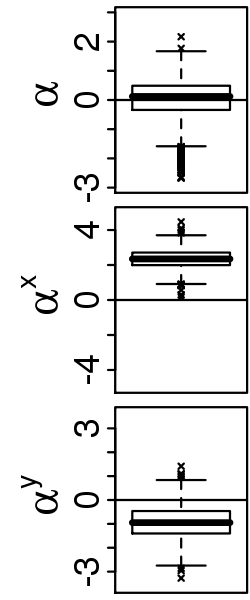}
\includegraphics[width=0.12\linewidth]{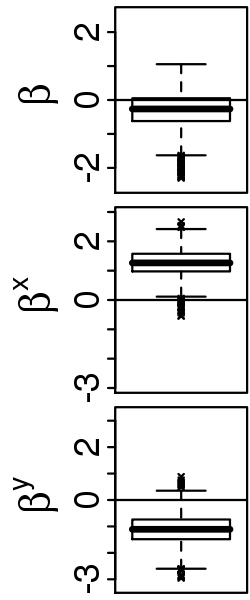}
\includegraphics[width=0.12\linewidth]{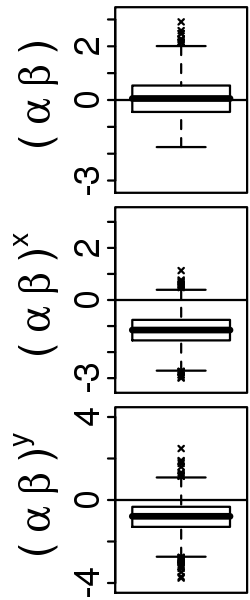}
\\Experiment 1 & & & Experiment 2
\end{tabular}
\end{center}
\caption{In experiment 1 (left), the method finds a disease effect $\alpha$ and a treatment effect $\beta$
  shared between the two views, plasma ($\mathbf{x}$) and lung ($\mathbf{y}$) tissues. In experiment 2 (right), only view-specific effects are found for plasma ($\mathbf{x}$) when integrating with the heart tissue ($\mathbf{y}$). No effects are found in heart.
The boxplots show quartiles and 95\% intervals of posterior mass of the effects, a consistently non-zero posterior distribution implies an effect found.
}
\label{realdata1}
\end{figure}

Blood plasma (168 lipids) and lung tissue (68 lipids) were integrated
with the method. The optimal number of clusters for plasma was 6 and
for lung 5, found by predictive likelihood. The method finds a disease
effect $\alpha$ and treatment effect $\beta$ shared by both views
(Fig.~\ref{realdata1}). The effect can be traced back to the
metabolite groups, by first identifying the responsible row of
$\mathbf{W}^x$ and hence component of $\mathbf{x}^{lat}$, and then the
metabolite cluster from the $\mathbf{V}^x$ corresponding to the
$\mathbf{x}^{lat}$ component.

The results imply that a cluster of 12 lipids in lung and a cluster of
20 lipids in blood are mutually coherently up-regulated due to
disease, and additionally up-regulated by the treatment. Another
cluster of 13 lipids in lung was found down-regulated due to
the disease and additionally down-regulated due to treatment. The
lipids of the down-regulated cluster are thus negatively correlated
with the up-regulated clusters. The results show that since no
consistent interaction term $(\alpha\beta)$ is found, there is no
indication that the treatment would cure the cancer effects. This
confirms our prior fear that the specific treatment might not be
efficient. The treatment does, however, affect the same groups of
lipids as the disease, so investigating it as a potential cure was not
a far-fetched hypothesis.

The up-regulated cluster of blood plasma contains abundant
triglycerides known to be coregulated, the up-regulated cluster of
lung contains lipotoxic ceramides \cite{Summers06} and proinflammatory
lysophosphatidylcholines \cite{Mehta05}, while the down-regulated
cluster of lung contains ether lipids, known as endogenous
antioxidants \cite{Brites04}. Our analysis reveals that the drug
treatment enhances, not diminishes, the proinflammatory lipid profile
found in the disease.

\subsubsection{Experiment 2. When connected to non-diseased tissue, only view-specific effects are found}

We then integrate plasma $\mathbf{x}$ with another tissue, heart (58 lipids) $\mathbf{y}$. The results (
Figure~\ref{realdata1}), show that the disease effect and the treatment
effect are found only in the view-specific component of plasma. This
implies that there are no shared effects between plasma and heart,
and in fact no consistent effects are found for the heart tissue. The
method finds, however, the same effects, $\alpha$ and $\beta$ in the
plasma tissue as in Experiment 1 and for the same cluster of
lipids, which is a sign that the method works well.
\section{Discussion}
We have generalized ANOVA-type multi-way analysis to cases where
multiple views of samples having a multi-way experimental setup are
available. The problem is solved by a hierarchical latent variable model that extends
the generative model of bayesian CCA to model multi-way covariate
information of samples by having population-specific priors on the
shared latent variable of CCA. Furthermore, the method is able
to decompose the covariate effects to shared and view-specific
effects, treating the multiple views as one covariate. Finally, the method is designed for cases with
high dimensionality and small sample-size, common in bioinformatics
applications. The small sample-size problem was solved by assuming that the variables come in correlated groups, which is reasonable for the metabolomics application. 

The modelling task is extremely difficult due to the complexity of the task and small sample-size. Hence it was striking that the method was capable of finding covariate effects in a real-world lipidomic multi-view, multi-way dataset.

In this work it was possible to estimate only three components, because the number of samples was extremely low: a shared
component and two view-specific components. If more than one
shared components are to be estimated, an unidentifiability problem
occurs, since there is a rotational ambiguity within the solution subspace. The
problem can be solved by a deflation-type method, where the components are computed one by one. Each
posterior sample is now considered as a converged starting point, and
a second component is added and the model is sampled with having the
first component fixed. The last sample of each new sampled chain is
collected for inference. 

\newpage
\bibliographystyle{unsrt}
\bibliography{arxiv09huopaniemi}

\end{document}